\pdfoutput=1

\documentclass[11pt]{article}

\usepackage[final]{acl}

\usepackage{times}
\usepackage{latexsym}
\usepackage{booktabs}
\usepackage{amssymb}
\usepackage{multirow}
\usepackage{graphicx}
\usepackage[T1]{fontenc}
\usepackage[utf8]{inputenc}

\usepackage{microtype}
\usepackage{colortbl}
\usepackage{pgfplots}
\usepackage{amsmath}
\usepackage{tabularx}
\usepackage{pifont}

\newcommand{\sfm}{SFM}
\newcommand{\sfms}{SFMs}

\newcommand{\fig}{Fig.}
\newcommand{\figs}{Figs.}
\newcommand{\tab}{Tab.}
\newcommand{\tabs}{Tabs.}
\newcommand{\sect}{Sec.}

\usepackage{inconsolata}

\definecolor{gradientstart}{RGB}{255,255,237} %
\definecolor{gradientend}{RGB}{0, 150, 255}      %
\newcommand{\maxvaluesa}{55.0}
\newcommand{\minvaluesa}{35.0}
\newcommand{\cellgradsa}[1]{%
  \pgfmathsetmacro{\colorgrad}{int((\maxvaluesa-#1)/(\maxvaluesa-\minvaluesa)*100)}%
  \edef\temp{\noexpand\cellcolor{gradientstart!\colorgrad!gradientend}%
              \noexpand\pgfmathprintnumber[fixed, precision=1, fixed zerofill]{#1}}\temp
}
\newcommand{\maxvaluesb}{49.0}
\newcommand{\minvaluesb}{9.0}
\newcommand{\cellgradsb}[1]{%
  \pgfmathsetmacro{\colorgrad}{int((\maxvaluesb-#1)/(\maxvaluesb-\minvaluesb)*100)}%
  \edef\temp{\noexpand\cellcolor{gradientend!\colorgrad!gradientstart}%
              \noexpand\pgfmathprintnumber[fixed, precision=1, fixed zerofill]{#1}}\temp
}
\newcommand{\maxvaluesc}{90.0}
\newcommand{\minvaluesc}{60.0}
\newcommand{\cellgradsc}[1]{%
  \pgfmathsetmacro{\colorgrad}{int((\maxvaluesc-#1)/(\maxvaluesc-\minvaluesc)*100)}%
  \edef\temp{\noexpand\cellcolor{gradientstart!\colorgrad!gradientend}%
              \noexpand\pgfmathprintnumber[fixed, precision=1, fixed zerofill]{#1}}\temp
}
\newcommand{\maxvaluesd}{75.0}
\newcommand{\minvaluesd}{45.0}
\newcommand{\cellgradsd}[1]{%
  \pgfmathsetmacro{\colorgrad}{int((\maxvaluesd-#1)/(\maxvaluesd-\minvaluesd)*100)}%
  \edef\temp{\noexpand\cellcolor{gradientstart!\colorgrad!gradientend}%
              \noexpand\pgfmathprintnumber[fixed, precision=1, fixed zerofill]{#1}}\temp
}
\newcommand{\maxvaluese}{40.0}
\newcommand{\minvaluese}{9.0}
\newcommand{\cellgradse}[1]{%
  \pgfmathsetmacro{\colorgrad}{int((\maxvaluese-#1)/(\maxvaluese-\minvaluese)*100)}%
  \edef\temp{\noexpand\cellcolor{gradientend!\colorgrad!gradientstart}%
              \noexpand\pgfmathprintnumber[fixed, precision=1, fixed zerofill]{#1}}\temp
}
\newcommand{\maxvaluesf}{75.0}
\newcommand{\minvaluesf}{45.0}
\newcommand{\cellgradsf}[1]{%
  \pgfmathsetmacro{\colorgrad}{int((\maxvaluesf-#1)/(\maxvaluesf-\minvaluesf)*100)}%
  \edef\temp{\noexpand\cellcolor{gradientstart!\colorgrad!gradientend}%
              \noexpand\pgfmathprintnumber[fixed, precision=1, fixed zerofill]{#1}}\temp
}
\newcommand{\maxvaluesg}{26.0}
\newcommand{\minvaluesg}{0.0}
\newcommand{\cellgradsg}[1]{%
  \pgfmathsetmacro{\colorgrad}{int((\maxvaluesg-#1)/(\maxvaluesg-\minvaluesg)*100)}%
  \edef\temp{\noexpand\cellcolor{gradientstart!\colorgrad!gradientend}%
              \noexpand\pgfmathprintnumber[fixed, precision=1, fixed zerofill]{#1}}\temp
}
\newcommand{\maxvaluesh}{17.0}
\newcommand{\minvaluesh}{15.0}
\newcommand{\cellgradsh}[1]{%
  \pgfmathsetmacro{\colorgrad}{int((\maxvaluesh-#1)/(\maxvaluesh-\minvaluesh)*100)}%
  \edef\temp{\noexpand\cellcolor{gradientstart!\colorgrad!gradientend}%
              \noexpand\pgfmathprintnumber[fixed, precision=1, fixed zerofill]{#1}}\temp
}
\newcommand{\maxvaluesi}{84.0}
\newcommand{\minvaluesi}{82.0}
\newcommand{\cellgradsi}[1]{%
  \pgfmathsetmacro{\colorgrad}{int((\maxvaluesi-#1)/(\maxvaluesi-\minvaluesi)*100)}%
  \edef\temp{\noexpand\cellcolor{gradientstart!\colorgrad!gradientend}%
              \noexpand\pgfmathprintnumber[fixed, precision=1, fixed zerofill]{#1}}\temp
}
\newcommand{\maxvaluesj}{77.0}
\newcommand{\minvaluesj}{45.0}
\newcommand{\cellgradsj}[1]{%
  \pgfmathsetmacro{\colorgrad}{int((\maxvaluesj-#1)/(\maxvaluesj-\minvaluesj)*100)}%
  \edef\temp{\noexpand\cellcolor{gradientstart!\colorgrad!gradientend}%
              \noexpand\pgfmathprintnumber[fixed, precision=1, fixed zerofill]{#1}}\temp
}
\newcommand{\xmark}{\ding{55}}%

\title{On the Evaluation of Speech Foundation Models\\ for Spoken Language Understanding}

\author{Siddhant Arora$^{1}$, Ankita Pasad$^{2}$, Chung-Ming Chien$^{2}$, Jionghao Han$^{1}$,  \\ {\bf Roshan Sharma$^{1}$, Jee-weon Jung$^{1}$, Hira Dhamyal$^{1}$, William Chen$^{1}$,}\\ {\bf Suwon Shon$^{3}$, Hung-yi Lee$^{4}$, Karen Livescu$^{2}$, Shinji Watanabe$^{1}$}\\
$^{1}$ Carnegie Mellon University, USA\\ 
$^{2}$ Toyota Technological Institute at Chicago\\
$^{3}$ ASAPP $^{4}$ National Taiwan University\\
  \texttt{\{siddhana\}@cs.cmu.edu} \\
}

\pgfplotsset{compat=1.18}
\begin{document}
\maketitle
\begin{abstract}
The Spoken Language Understanding Evaluation (SLUE) suite of benchmark tasks was recently introduced to address the need for open resources and benchmarking of complex spoken language understanding (SLU) tasks, including both classification and sequence generation tasks, on natural speech. The benchmark has demonstrated preliminary success in using pre-trained speech foundation models (\sfm) for these SLU tasks. However, the community still lacks a fine-grained understanding of the comparative utility of different \sfms.
Inspired by this, we ask: which \sfms\space offer the most benefits for these complex SLU tasks, and what is the most effective approach for incorporating these \sfms?
To answer this, 
we 
perform an extensive evaluation of multiple supervised and self-supervised \sfms\space using several evaluation protocols: (i) \emph{frozen} \sfms\space with a \emph{lightweight} prediction head, (ii) \emph{frozen} \sfms\space with a \emph{complex} prediction head, and (iii) \emph{fine-tuned} \sfms\space with a \emph{lightweight} prediction head.
Although the supervised  \sfms\space are pre-trained on much more speech recognition data (with labels), they do not always outperform self-supervised \sfms; the latter tend to perform at least as well as, and sometimes better than, supervised \sfms, especially on the sequence generation tasks in SLUE.
While there is no \emph{universally} optimal way of incorporating \sfms, the \emph{complex} prediction head gives the best performance for most tasks, although it increases the inference time.
We also introduce an open-source toolkit and performance leaderboard, SLUE-PERB, for these tasks and modeling strategies.

\end{abstract}

\section{Introduction}
Spoken language understanding (SLU) refers to tasks that require extracting semantics from spoken utterances. 
SLU systems have important applications, for example, in voice assistants and conversational agents, and have attracted increasing interest in recent years~\cite{socialbot, snips-voice-platform}.
SLU encompasses a wide range of tasks, such as predicting intents and slots~\cite{Lugosch_FSC,SLURP,coucke2018snips}, recognizing entity mentions and labels~\cite{SLURP, earnings21}, detecting the speaker's sentiment~\cite{iemocap} and modeling the topic of a spoken dialogue~\cite{SWB_DA_res,SWB}. More recently, there has been significant interest in tackling more complex tasks like question answering~\cite{li2018spoken, shon2023slue} or summarization \cite{sharmaEnd}. 

The Spoken Language Understanding Evaluation (SLUE)~\cite{shon2022slue,shon2023slue} suite of benchmark tasks was recently proposed to address the lack of sufficiently complex and varied tasks on natural (rather than synthetic or read) speech from public datasets.
SLUE uses annotated natural speech from conversations and monologues and includes both classification and sequence generation tasks. 
Traditional SLU models use a pipeline~\cite{palmer-ostendorf-2001-improving,Horlock2003DiscriminativeMF,BECHET2004207} of an automatic speech recognition (ASR) system followed by a natural language understanding (NLU) system. 
End-to-end (E2E) SLU systems~\cite{ESPnet-SLU,end-to-end1} have also been explored to mitigate the impact of error propagation observed in pipeline approaches and take advantage of the information in the audio signal beyond the word content.  

A recent trend in E2E models has been the use of pre-trained speech foundation models (\sfm)~\cite{mohamed2022self, wavlm,HuBERT,whisper,owsm} that can learn useful representations for a large number of tasks. 
Due to the increasing diversity of models,
benchmarks are important to compare the performance of \sfms\space on multiple downstream tasks. 
Performance benchmarks like SUPERB (Speech processing Universal PERformance Benchmark)~\cite{SUPERB} have facilitated standardized comparison of pre-trained \sfms\space across a diverse range of speech-processing tasks. 
However, such benchmarks lack coverage of challenging and realistic SLU tasks. Hence, the community lacks a fine-grained understanding of the relative merits of different \sfms\space and different ways to use them for downstream SLU tasks.

Motivated by these shortcomings, we introduce \emph{SLUE-PERB} (Spoken Language Understanding Evaluation PERformance Benchmark), specifically designed to evaluate representations extracted from pre-trained \sfms\space on complex SLU tasks. 
We use this benchmark to answer two main questions: (i) which \sfms\space
are most useful for these tasks, and (ii) how do different ways of using these \sfms, varying in their compute budget, compare. Our study addresses various questions concerning SLU systems, such as whether supervised \sfms\space are more beneficial than self-supervised \sfms, whether \sfms\space are effective as frozen feature extractors or should be fine-tuned on downstream tasks, and whether the complexity of prediction heads affects the performance trends.

We conduct a comprehensive analysis by examining three types of \sfms: (i) \textit{self-supervised} learning (SSL) speech models~\cite{wav2vec2,HuBERT,wavlm} trained on unlabeled speech data; (ii) (weakly) \textit{supervised} ASR (and speech translation) models~\cite{whisper,owsm} pre-trained on large labeled corpora; and (iii) \textit{supervised} SLU models pre-trained on external SLU corpora~\cite{swbd_sentiment,SLURP}. Our extensive experiments are performed on the SLUE benchmark~\cite{shon2022slue,shon2023slue}, which provides curated data for Sentiment Analysis (SA), Named Entity Recognition (NER), Named Entity Localization (NEL), Dialogue Act Classification (DAC), Question Answering (QA) and Summarization (SUMM).
\begin{table*}[t]
  \centering
  \resizebox{\linewidth}{!}{
  \begin{tabular}{llccclll}
    \toprule
    \multirow{2}{*}{Dataset} & \multirow{2}{*}{Speaking Style}  &\multicolumn{3}{c}{Size (hours)} & \multirow{2}{*}{Tasks}  & \multirow{2}{*}{Output} & \multirow{2}{*}{Metric}\\\cmidrule(lr){3-5}
    & &  Train & Dev & Test & & &\\
    \midrule
    \multirow{2}{*}{SLUE-VoxCeleb} & \multirow{2}{*}{Conversational} & \multirow{2}{*}{\hphantom{0}12.8} & \multirow{2}{*}{\hphantom{0}2.1} & \multirow{2}{*}{\hphantom{0}9.0} & SA$^*$  & sentiment class & F1\\
    &  & & & & ASR$^\dagger$ & text transcript & WER\\\cmidrule(lr){6-8}
    \multirow{3}{*}{SLUE-VoxPopuli} & \multirow{3}{*}{Orated Speech}  & \multirow{3}{*}{\hphantom{0}14.5} & \multirow{3}{*}{\hphantom{0}5.0} & \multirow{3}{*}{\hphantom{0}4.9} & NER$^\dagger$  & (entity phrase, entity tag) pairs & Label F1, F1\\
    &  & & & & NEL$^\S$ & (entity start time, entity end time) pairs & Frame F1\\
    &  & & & & ASR$^\dagger$ & text transcript & WER\\
    \cmidrule(lr){6-8}
    SLUE-HVB & Scripted conversation & \hphantom{00}6.8 & \hphantom{0}1.0 & \hphantom{0}3.6 & DAC$^*$ & dialogue act classes & F1\\
    \cmidrule(lr){6-8}
    SLUE-SQA-5 & Read speech & 244.0 & 21.2 & 25.8 & QA$^\S$ & (answer start time, answer end time) & Frame F1\\
    \cmidrule(lr){6-8}
    SLUE-TED & Orated Speech & 664.0 & 81.0 & 84.0 & SUMM$^\dagger$ & text summary & ROUGE-L, \\
    & & & & & & & BERTScore\\
    \bottomrule
    \multicolumn{7}{l}{$^*$: Classification, $^\dagger$: Sequence generation, $^\S$: Temporal Alignment }
  \end{tabular}
  }
  \vspace{-0.2cm}
    \caption{Overview of the datasets~\cite{shon2022slue,shon2023slue} and tasks in SLUE-PERB.  "WER" = "word error rate."}
  \label{tab:slue-datasets}
  \vspace{-0.5cm}
\end{table*}
The key contributions are:
\begin{itemize}
    \item We compare representations extracted from various pre-trained \sfms\space across all SLUE tasks. Our experiments reveal that pre-trained ASR \sfms\space excel in classification tasks, while SSL \sfms\space either outperform or perform comparably to supervised ASR \sfms\space in sequence generation tasks.
    \item We evaluate different modeling strategies and find that the performance improves, and the performance gap between different \sfms\space reduces, as we increase the prediction head size or fine-tune the pre-trained \sfms\space instead of using frozen representations.
    
    \item While no single method is \emph{universally} optimal for all tasks, employing a complex prediction head is the best performing strategy for most tasks when inference speed is not a limiting factor. On the other hand, fine-tuned \sfms\space with a lightweight prediction head are a good option if latency is a concern.
    \item We release our code publicly so that researchers can easily reproduce our results and test their own pre-trained \sfms.
\end{itemize}

\section{Related Work}

\subsection{Pre-trained speech foundation models}
The earliest self-supervised speech model, pre-trained on large amounts of unlabeled data, to show improvements in large-scale ASR was wav2vec~\cite{schneider2019wav2vec}. Since then, the community has developed a variety of pre-trained self-supervised \sfms~\cite{mohamed2022self} and their representations have been successfully incorporated into task-specific models spanning many applications.

Recently, supervised \sfms \ pre-trained on large amounts of paired or weakly paired speech-text data have gained in popularity. Studies ~\cite{UniverSLU,arora2023study} have shown that these supervised \sfms \ can be fine-tuned to achieve state-of-the-art (SOTA) performance on certain downstream tasks. But it remains to be seen how supervised pre-training compares with self-supervised \sfms \ on complex language understanding tasks like those in SLUE. 

The few studies so far on \sfms \ for SLU~\cite{SUPERB, shon2022slue, shon2023slue, wav2seq, chien2023few, chou2023toward} focus on only selected SLU tasks, a single pre-trained \sfm, or simpler SLU tasks. With SLUE-PERB, we aim to fill this knowledge gap by studying the applicability of different types of \sfms \ and modeling strategies on a variety of SLU tasks.

\subsection{Performance benchmarks} Performance benchmarks have been widely used to study performance on downstream tasks and the information encoded in \sfms. Among them, \emph{SUPERB}~\cite{SUPERB} is a popular benchmark developed for SSL \sfms. It includes a variety of downstream tasks from speech recognition, speaker recognition, emotion recognition, to simple SLU tasks like intent classification and slot filling. It uses a shared evaluation protocol, combining a frozen \sfm\space with a lightweight prediction head for each task. %
Extensions of the benchmark to different languages (LeBenchmark, IndicSUPERB, ML-SUPERB~\cite{parcollet2023lebenchmark,IndicSUPERB,shi23g_interspeech}), modalities (AV-SUPERB~\cite{tseng2023avsuperb}), and tasks (SUPERB-SG~\cite{tsai2022superb}) have been proposed. 

Though such benchmarks have tremendous value, they lack coverage of challenging and practical SLU tasks. 
Motivated by this, SLUE~\cite{shon2022slue,shon2023slue} was proposed to focus on more challenging SLU tasks on freely available annotated natural speech datasets, including conversational or long-discourse speech, as shown in \tab~\ref{tab:slue-datasets}.
However, the original SLUE tasks do not have a standardized evaluation protocol with an interface to a benchmark.
Additionally, SLUE primarily aimed to compare various pipeline and E2E SLU systems rather than analyze the comparative efficacy of different SFMs.
To address these issues, we introduce SLUE-PERB, which exhaustively evaluates various pre-trained \sfms \ across different evaluation settings on these complex SLU tasks.

\section{The SLUE-PERB benchmark}
\label{sec:performance_benchmark}
SLUE-PERB is an open-source testbed for evaluating \sfms \ on SLU tasks.
\begin{table*}[t]
  \centering
  \resizebox{\linewidth}{!}
  {
  \begin{tabular}{ccclp{37mm}p{28mm}}
    \toprule
    Type & Speech Foundation Model & Architecture & Model size & Dataset (size in hours) & Objective\\
    \midrule
    \multirow{3}{*}{SSL}& Wav2Vec2 (large) (\citeauthor{wav2vec2}) & 7-Conv 24-Trans & 317.4M & LibriLight 60k (60k) & contrastive\\
    & HuBERT (large) (\citeauthor{HuBERT}) & 7-Conv 24-Trans & 316.6M & LibriLight 60k (60k) & masked prediction\\
    & WavLM (large) (\citeauthor{wavlm}) & 7-Conv 24-Trans & 315.5M & Mix 94k (94k) & masked prediction + de-noising\\ \midrule
    \multirow{2}{*}{ASR} & Whisper (med.) (\citeauthor{whisper}) & 2-Conv 24-Trans & 315.7M & Web data (680k) & ASR, ST\\
    & OWSM (3.1) (\citeauthor{peng2024owsm})& 2-Conv 18-Branch & 560.8M & Open-source ASR + ST data (180k) &  ASR, ST \\ \midrule
    \multirow{2}{*}{SLU} & SWBD Sentiment (\citeauthor{ESPnet-SLU}) & 2-Conv 12-Conf & \hphantom{0}82.2M & SWBD Sentiment (260) & SLU\\
    & SLURP (\citeauthor{ESPnet-SLU}) & 2-Conv 12-Conf & \hphantom{0}83.2M & SLURP (58) & SLU\\
    \bottomrule
  \end{tabular}
  }
  \caption{Summary of the \emph{encoder} of self-supervised and supervised pre-trained \sfms \ used in this work. 
  The Mix 94k dataset is a mixture of LibriLight 60k~\cite{librilight}, GigaSpeech 10k~\cite{gigaspeech}, and VoxPopuli 24k~\cite{voxpopuli}.
  }
  \label{tab:pre-trained-speech}
    \vspace{-0.5cm}
\end{table*}

\subsection{Tasks}
\label{subsec:toolkit_task}
Our benchmark currently focuses on the datasets from SLUE~\cite{shon2022slue} and SLUE Phase-2~\cite{shon2023slue}. We provide support for 6 SLUE tasks, shown in \tab~\ref{tab:slue-datasets}.
SA is an utterance-level classification task of identifying the sentiment of an utterance. NER is a sequence prediction task of detecting the named entities and labeling their tags in a spoken utterance. NEL involves locating the entities, i.e., predicting the start and end timestamps of any entity in the audio. DAC is an utterance-level multi-label, multi-class classification task that identifies the function(s) of an utterance in a spoken conversation, such as a statement, a question, etc. QA involves locating the answer (i.e. predicting the start and end timestamps) in a spoken document given a spoken question. SUMM is a sequence prediction task that involves generating a text summary of a long speech input.
\sect~\ref{subsec: dataset} in the Appendix provides additional dataset details.

\subsection{Pre-trained speech foundation models}
\label{subsec:model}
We experiment with the following three types of pre-trained \sfms, summarised in \tab~\ref{tab:pre-trained-speech}, with additional details in \sect~\ref{subsec: model_details} in the Appendix.

{\noindent \bf Self-supervised \sfms:}
To incorporate SSL \sfms, we follow prior work~\cite{SUPERB} and use a weighted sum of the hidden layer representations of SSL encoder to generate speech representations.

{\noindent \bf Supervised ASR \sfms:}
We use representations derived from the hidden layers of the encoder of supervised encoder-decoder ASR SFMs. The use of the encoder alone makes the comparisons with SSL-based encoders more straightforward, and also follows the practice of prior work using supervised ASR SFMs for other downstream tasks~\cite{gong2023whisper}. However, in future work, we plan to study the use of the pre-trained decoder as well.

{\noindent \bf Supervised SLU \sfms:}
Since most SLU tasks have limited labeled data, our benchmark also evaluates the impact of pre-training using an external SLU corpus. As in the case of supervised ASR models, we use the encoder of the pre-trained model to extract speech representations. For SLU SFMs, we choose pre-training SLU corpora designed for the same task as the target SLU data. Hence, we use SLU model pre-trained on the SWBD Sentiment dataset for the SA task and SLU model pre-trained on SLURP for all other tasks.
 
\subsection{Evaluation Protocols}
\label{subsec:protocol}
This section provides a high-level overview of the various prediction heads and approaches for leveraging \sfms \ investigated in this study. We consistently employ a learned weighted sum of hidden layers of \sfms \ to generate speech representations across all 3 approaches. Further details about the evaluation setup are in \sect~\ref{sec: experiment}.

{\noindent \bf Lightweight prediction head:}
We first experiment with using a similar evaluation protocol to SUPERB, where the pre-trained \sfm \ is kept frozen, with a lightweight prediction head learned on top of it to perform classification or sequence generation. 
Depending on the task, this lightweight prediction head usually consists of a classification layer or a shallow encoder with CTC.
As in SUPERB, we use weighted combinations of hidden layer activations as the input to the classifier or encoder. 
This evaluation protocol not only facilitates quick comparison of various \sfms \ but also promotes the development of models capable of performing well across multiple tasks without the need for task-specific fine-tuning.
Unlike SUPERB, SLUE-PERB does not restrict its evaluation solely to SSL \sfms. 

{\noindent \bf Fine-tuned representations:} 
Another popular paradigm for incorporating pre-trained \sfms \ is fine-tuning the \sfms \ along with a lightweight prediction head.
While there are multiple approaches to fine-tune \sfms, including parameter-efficient approaches like LoRA~\cite{LoRA}, full fine-tuning has been most commonly used in prior works~\cite{fairseq,shon2022slue}. However, this approach significantly increases the computation cost during fine-tuning, which might make it challenging to use in scenarios with a limited computation budget.

{\noindent \bf Complex prediction head:} Motivated by prior works~\cite{zaiem2023speech,zaiem2023speech2} that show a change in benchmark results with a change in prediction head architectures, we investigate increasing the complexity of the prediction head while keeping the \sfms \ frozen. In this protocol, we experiment with a ``prediction head'' based on an encoder-decoder architecture. The input to this prediction head is a sequence of pre-trained speech representations and the output is a sequence of text tokens denoting the SLU label sequence.
While this setting does increase inference time, it serves as a middle ground between the ``Lightweight prediction head'' and ``Fine-tuned representations'' settings in terms of the number of trainable parameters and has been used in prior works on SLU~\cite{ESPnet-SLU}.

\begin{table*}[t]
  \centering
    \resizebox {\linewidth} {!} {
\begin{tabular}{cc|cccccccccc}
\toprule
Evaluation & Pre-Trained &  \multicolumn{2}{c}{SLUE-VoxCeleb}  &  \multicolumn{4}{c}{SLUE-VoxPopuli} & SQA-5 & \multicolumn{2}{c}{SLUE-TED} & SLUE-HVB \\
\cmidrule(r){3-4}\cmidrule(r){5-8}\cmidrule(r){9-9}\cmidrule(r){10-11}\cmidrule(r){12-12}
Protocol &  Model & SA & ASR & \multicolumn{2}{c}{NER} & ASR & NEL & QA & \multicolumn{2}{c}{SUMM} & DAC\\
& & F1 $\uparrow$ & WER $\downarrow$ & Label F1	$\uparrow$ & F1 $\uparrow$ & WER $\downarrow$ & Frame F1 $\uparrow$ & Frame F1 $\uparrow$ & ROUGE-L $\uparrow$ & BERTScore $\uparrow$ & F1 $\uparrow$\\
\midrule
\multirow{2}{*}{Lightweight} & HuBERT (large) & \cellgradsa{41.0} & \cellgradsb{19.0} & \cellgradsc{76.5} & \cellgradsd{59.3} & \cellgradse{14.2} & \cellgradsf{67.7} & \cellgradsg{12.0} & \xmark & \xmark & \cellgradsj{48.0}\\
& Wav2Vec2 (large) & \cellgradsa{40.6} & \cellgradsb{21.7} & \cellgradsc{73.6} & \cellgradsd{57.5} & \cellgradse{16.0} & \cellgradsf{64.1} & \hphantom{0}\cellgradsg{6.0} & \xmark & \xmark & \cellgradsj{51.2}\\
\multirow{2}{*}{prediction}& WavLM (large) & \cellgradsa{43.3} & \cellgradsb{14.1} & \cellgradsc{80.6} & \cellgradsd{64.5} & \cellgradse{10.4} & \cellgradsf{72.0} & \cellgradsg{17.4} & \xmark & \xmark & \cellgradsj{54.6}\\
& Whisper (medium) & \cellgradsa{49.6} & \cellgradsb{15.0} & \cellgradsc{79.6} & \cellgradsd{63.1} & \cellgradse{12.5} & \cellgradsf{71.8} & \hphantom{0}\cellgradsg{0.1} & \xmark & \xmark & \cellgradsj{59.7}\\
\multirow{2}{*}{head} & OWSM (3.1) & \cellgradsa{47.2} & \cellgradsb{17.4} & \cellgradsc{78.4} & \cellgradsd{61.7} & \cellgradse{12.8} & \cellgradsf{70.5} & \cellgradsg{14.0} & \xmark & \xmark & \cellgradsj{66.3}\\
& Pre-trained SLU & \cellgradsa{36.4} & \cellgradsb{47.5} & \cellgradsc{60.8} & \cellgradsd{45.5} & \cellgradse{39.1} & \cellgradsf{47.8} & \hphantom{0}\cellgradsg{2.0} & \xmark & \xmark & \cellgradsj{54.4}\\\midrule
\multirow{2}{*}{Complex} & HuBERT (large) & \cellgradsa{52.2} & \cellgradsb{15.5} & \cellgradsc{78.5} & \cellgradsd{63.1} & \cellgradse{13.0} & \cellgradsf{69.8} & \cellgradsg{21.4} & \cellgradsh{16.0} & \cellgradsi{83.4} & \cellgradsj{66.1}\\
& Wav2Vec2 (large) & \cellgradsa{53.3} & \cellgradsb{17.2} & \cellgradsc{78.2} & \cellgradsd{63.7} & \cellgradse{14.0} & \cellgradsf{71.2} & \cellgradsg{18.8} & \cellgradsh{16.2} & \cellgradsi{83.0} & \cellgradsj{65.8}\\
\multirow{2}{*}{prediction}& WavLM (large) & \cellgradsa{52.0} & \cellgradsb{11.4} & \cellgradsc{82.7} & \cellgradsd{69.7} & \cellgradse{10.1} & \cellgradsf{72.6} & \cellgradsg{22.5} & \cellgradsh{16.4} & \cellgradsi{83.0} & \cellgradsj{67.4}\\
& Whisper (medium) & \cellgradsa{51.0} & \cellgradsb{14.9} & \cellgradsc{79.2} & \cellgradsd{64.1} & \cellgradse{13.2} & \cellgradsf{70.1} & \cellgradsg{1.6} & \cellgradsh{16.0} & \cellgradsi{83.8} & \cellgradsj{67.8}\\
\multirow{2}{*}{head}& OWSM (3.1) & \cellgradsa{52.8} & \cellgradsb{16.5} & \cellgradsc{79.6} & \cellgradsd{66.0} & \cellgradse{12.6} & \cellgradsf{68.6} & \cellgradsg{20.3} & \cellgradsh{16.5} & \cellgradsi{83.6} & \cellgradsj{69.4}\\
& Pre-trained SLU & \cellgradsa{49.7} & \cellgradsb{36.4} & \cellgradsc{68.7} & \cellgradsd{54.8} & \cellgradse{28.5} & \cellgradsf{54.4} & \hphantom{0}\cellgradsg{3.2} & \cellgradsh{15.4} & \cellgradsi{82.9} & \cellgradsj{66.3}\\
\midrule
\multirow{3}{*}{Fine-tuning} & HuBERT (large) & \cellgradsa{46.5} & \cellgradsb{14.8} & \cellgradsc{78.8} & \cellgradsd{62.6} & \cellgradse{12.0} & \cellgradsf{69.4} & \xmark & \xmark & \xmark & \cellgradsj{72.7}\\
& Wav2Vec2 (large) & \cellgradsa{45.0} & \cellgradsb{14.7} & \cellgradsc{78.2} & \cellgradsd{62.9} & \cellgradse{11.7} & \cellgradsf{68.6} & \xmark & \xmark & \xmark & \cellgradsj{71.3}\\
& WavLM (large) & \cellgradsa{47.9} & \cellgradsb{12.1} & \cellgradsc{82.5} & \cellgradsd{66.3} & \hphantom{0}\cellgradse{9.7} & \cellgradsf{71.7} & \xmark & \xmark & \xmark & \cellgradsj{71.5}\\
\multirow{3}{*}{representations}& Whisper (medium) & \cellgradsa{51.8} & \cellgradsb{20.5} & \cellgradsc{76.9} & \cellgradsd{59.8} & \cellgradse{18.2} & \cellgradsf{56.6} & \xmark & \xmark & \xmark & \cellgradsj{69.8}\\
& OWSM (3.1) & \cellgradsa{47.8} & \cellgradsb{15.0} & \cellgradsc{78.5} & \cellgradsd{61.5} & \cellgradse{14.3} & \cellgradsf{65.1} & \xmark & \xmark & \xmark & \cellgradsj{72.1}\\
& Pre-trained SLU & \cellgradsa{46.1} & \cellgradsb{34.6} & \cellgradsc{60.8} & \cellgradsd{47.6} & \cellgradse{37.1} & \cellgradsf{49.1} & \xmark & \xmark & \xmark & \cellgradsj{68.7}\\
\bottomrule
\end{tabular}
}
\vspace{-0.1in}
  \caption{Performance of various SSL, supervised ASR, and SLU representations on the test set of SLUE tasks using various evaluation protocols in SLUE-PERB. The symbol \xmark\ indicates that the results were not computed, either due to the inability to perform summarization without a decoder or because fine-tuning representations on SQA-5 and SLUE-TED corpora is not feasible within our computational budget. } 
\label{tab:main-test-results}
\vspace{-0.1in}
\end{table*}

\section{Experiments}
\label{sec: experiment}
We conduct our analysis by examining various \sfms \ as introduced in \tab~\ref{tab:pre-trained-speech}.
Training hyper-parameters are selected based on validation performance. More details can be found in \sect~\ref{subsec:appendix_exp} in the Appendix. All our models and config files will be publicly available upon acceptance of the paper.

{\noindent \bf Lightweight prediction head:}
For the SA task, we mean-pool the extracted features from the \sfms\space across time, and then pass the pooled representation through a linear layer to compute the probability for each sentiment class. The lightweight classification layers are trained using cross-entropy loss. In the case of DAC, we follow a similar procedure of mean-pooling followed by a linear layer. As this is a multi-label classification task, we use a sigmoid activation to compute the probability for each dialogue class and train the linear layer using binary cross entropy loss. During inference, classes with a probability greater than 0.5 are considered positive.

For sequence prediction and temporal alignment tasks like ASR, NER, NEL and QA, we pass the extracted features through a shallow encoder trained with CTC loss. NER and ASR models use a 2-layer conformer encoder as the prediction head and follow a similar input-output formulation as in~\citet{pretrain_slue}. For NEL, following~\citet{shon2023slue}, we perform greedy CTC decoding on the NER model to obtain frame-level alignments, which are used to get entity start and end time-stamps.
For the QA task, the input to the model is the concatenation of the question and document audio, and the output is the concatenation of the question and document transcript where the answer is delimited by a special character (See \sect~\ref{subsec:appendix_exp}).
Since QA involves more complex language understanding, we use a 4-layer conformer encoder\footnote{2-layer conformer encoder achieved poor performance} and again get timestamps using greedy CTC decoding. 
We experimented with encoder-only CTC training for SUMM as well but found that coherent summaries cannot be produced without a decoder and, hence, we do not report results with a lightweight prediction head for SUMM.

{\noindent \bf Complex prediction head:}
The complex prediction head is an encoder-decoder architecture consisting of a 12-layer conformer encoder and a 6-layer transformer decoder, which takes as input the weighted sum of representations from pre-trained speech models and outputs the SLU label sequence.
For classification tasks, the SLU label sequence comprises the ASR transcript concatenated after the SLU class label, following prior work~\cite{ESPnet-SLU}. The SLU label sequences for sequence generation and temporal alignment tasks are identical to those in the "lightweight prediction head".  For the SUMM task, the TED talks are too long to fit in a GPU, and prior work~\cite{espnet_summ} has shown that very little performance is lost by using only the first 30 seconds of input audio in the SLUE-TED dataset.\footnote{This may be partly attributed to the dataset characteristics, where both the audio and ground truth summaries are sourced from TED talks. Upon manual inspection, we observe that the summaries often serve as an introduction to the talk, and the key information in the talk summary is often found within the first 30 seconds.} Hence, we truncate all the audios to 30 seconds since the TED talks were too long to fit in a GPU. Since we experiment with various \sfms\ using the same setup (i.e. using only 30-second input), we believe it is a fair comparison for gaining insights into the relative utility of various \sfms. Our approach can be extended to use more than 30 seconds of input by developing additional strategies to deal with long-form inputs, which will be an interesting future direction.
We follow prior works~\cite{shon2023slue} to first pre-train the model for ASR on the TEDLIUM-3 corpus, and then train the model for summarization on the SLUE-TED dataset.

{\noindent \bf Fine-tuned representations:}
 The prediction head architecture and model inputs/outputs are identical to those of the "lightweight prediction head" setup for all the tasks. We omit the QA and SUMM tasks in this setting, as fine-tuning representations on the SQA-5 and SLUE-TED corpora is too computationally expensive.
\begin{figure}[t]
\centering
\includegraphics[width=0.96\linewidth]{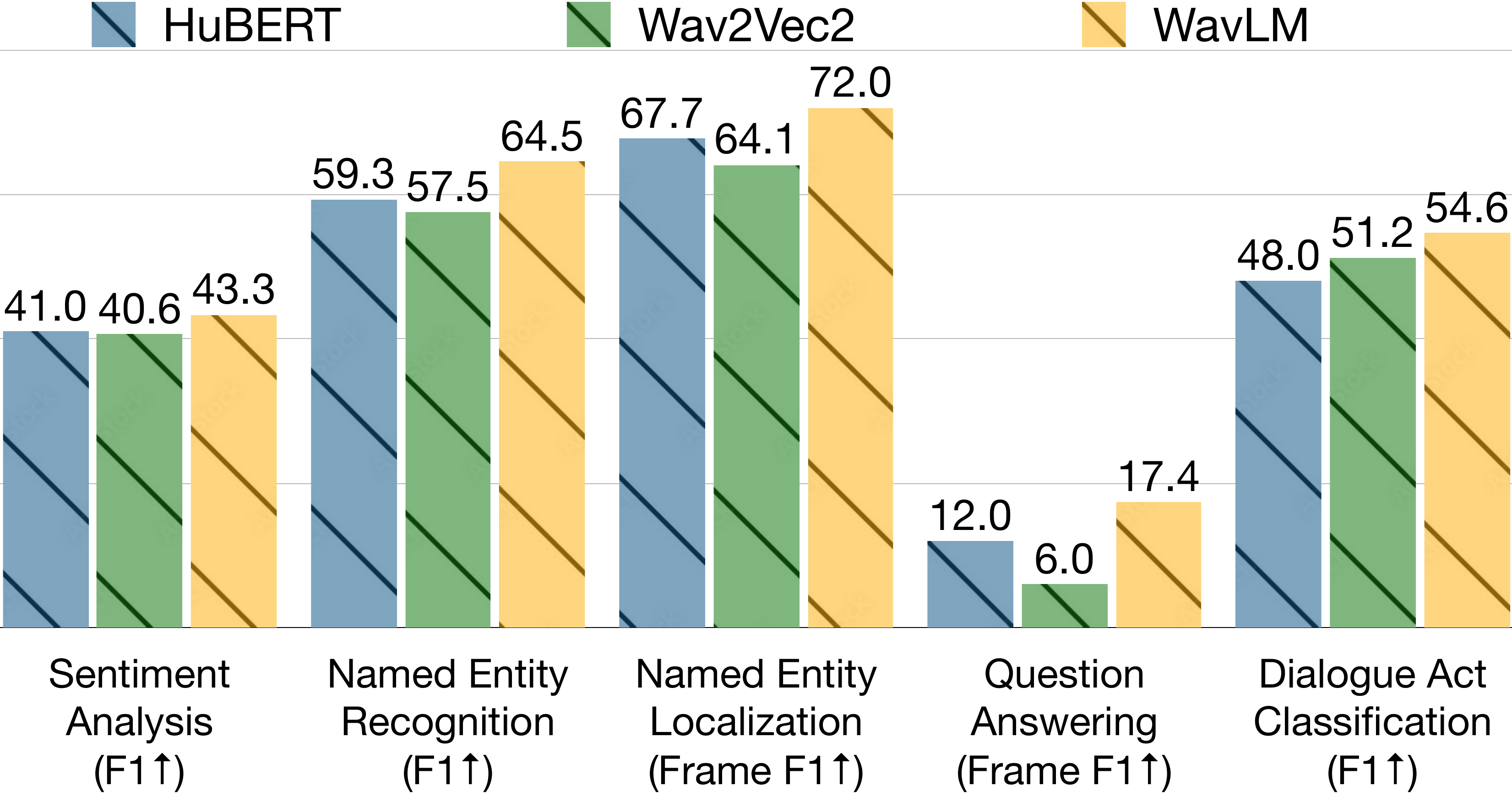}
\vspace{-0.1in}
\caption{Performance of various SSL \sfms\space with a lightweight prediction head on SLUE tasks.}
\label{fig:SSL_simple}
\vspace{-0.1in}
\end{figure}
\begin{figure}[t]
\centering
\includegraphics[width=0.9\linewidth]{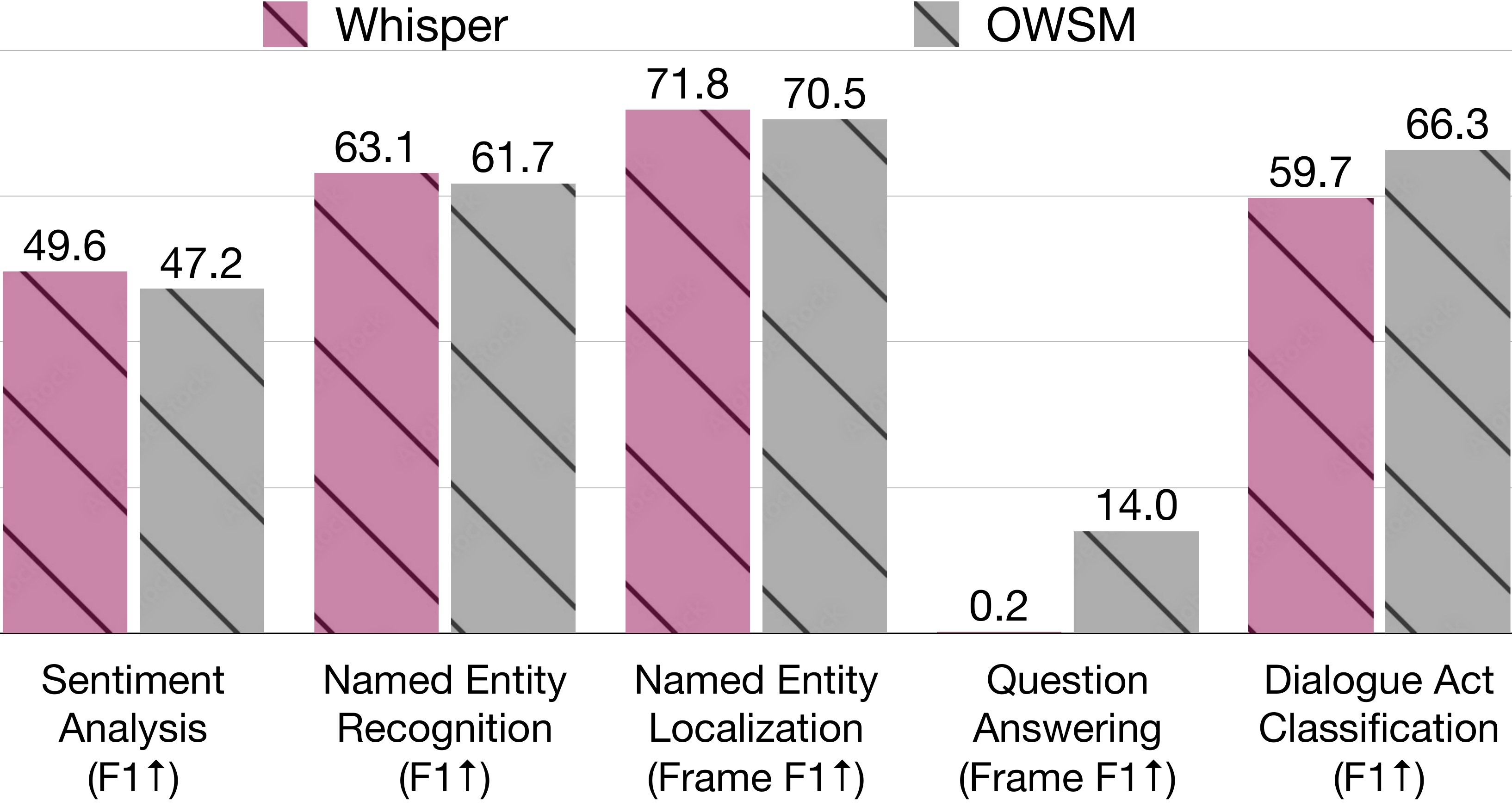}
\vspace{-0.1in}
\caption{Performance of various supervised ASR \sfms\space with a lightweight prediction head on SLUE tasks.
}
\label{fig:ASR_simple}
\vspace{-0.2in}
\end{figure}

\section{Results}
\begin{figure}[t]
\centering
\includegraphics[width=0.98\linewidth]{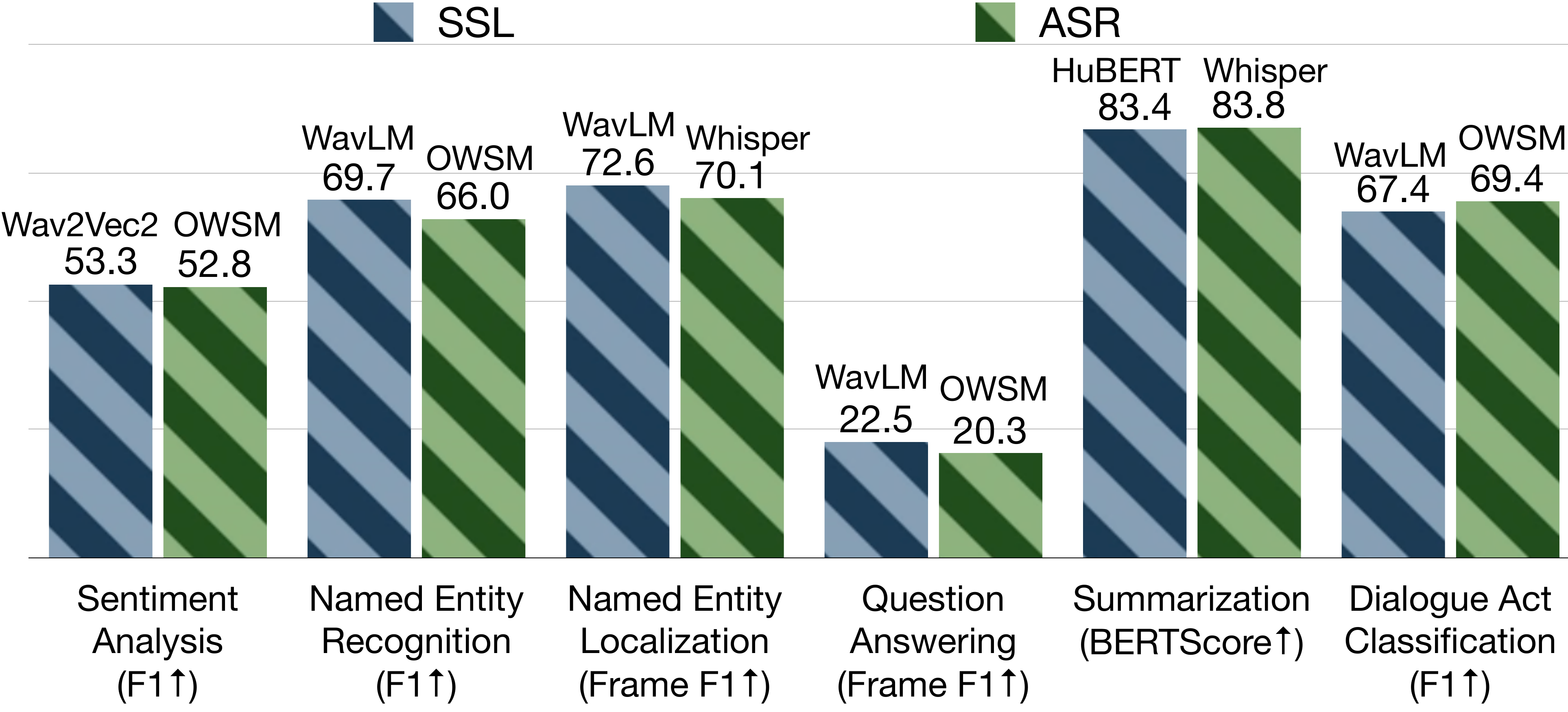}
\vspace{-0.1in}
\caption{Performance of best performing SSL and ASR \sfms\space with complex prediction head on SLUE tasks. The label for each bar is the specific \sfm\space chosen.}
\label{fig:Full_complex}
\vspace{-0.1in}
\end{figure}

\begin{figure}[t]
\centering
\includegraphics[width=0.98\linewidth]{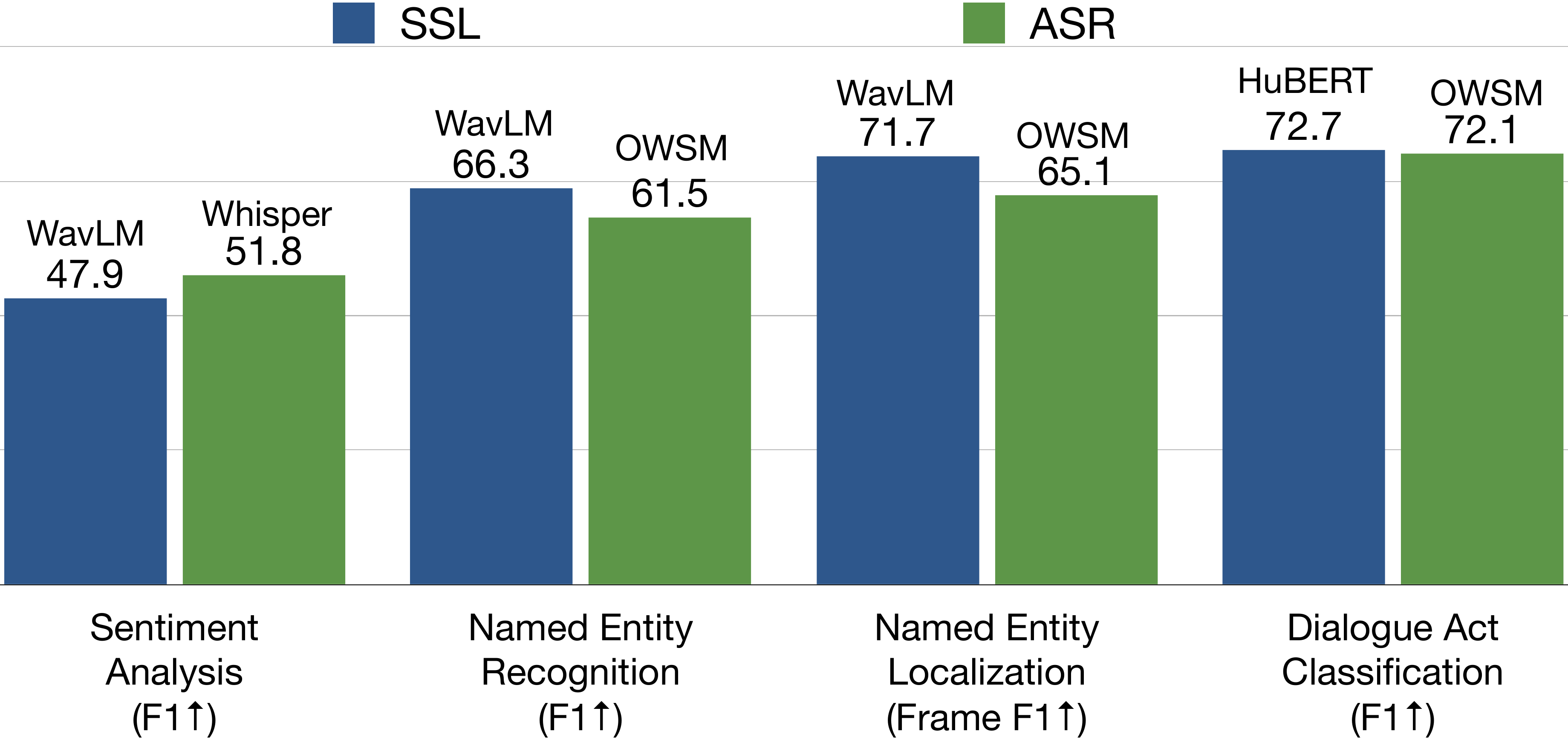}
\vspace{-0.1in}
\caption{Performance of best performing SSL and ASR \sfms\space with fine-tuned representations on SLUE tasks. The label for each bar is the specific \sfm\space chosen.}
\label{fig:Full_fine-tune}
\vspace{-0.1in}
\end{figure}

\begin{figure*}[t]
\centering
\includegraphics[width=0.75\textwidth]{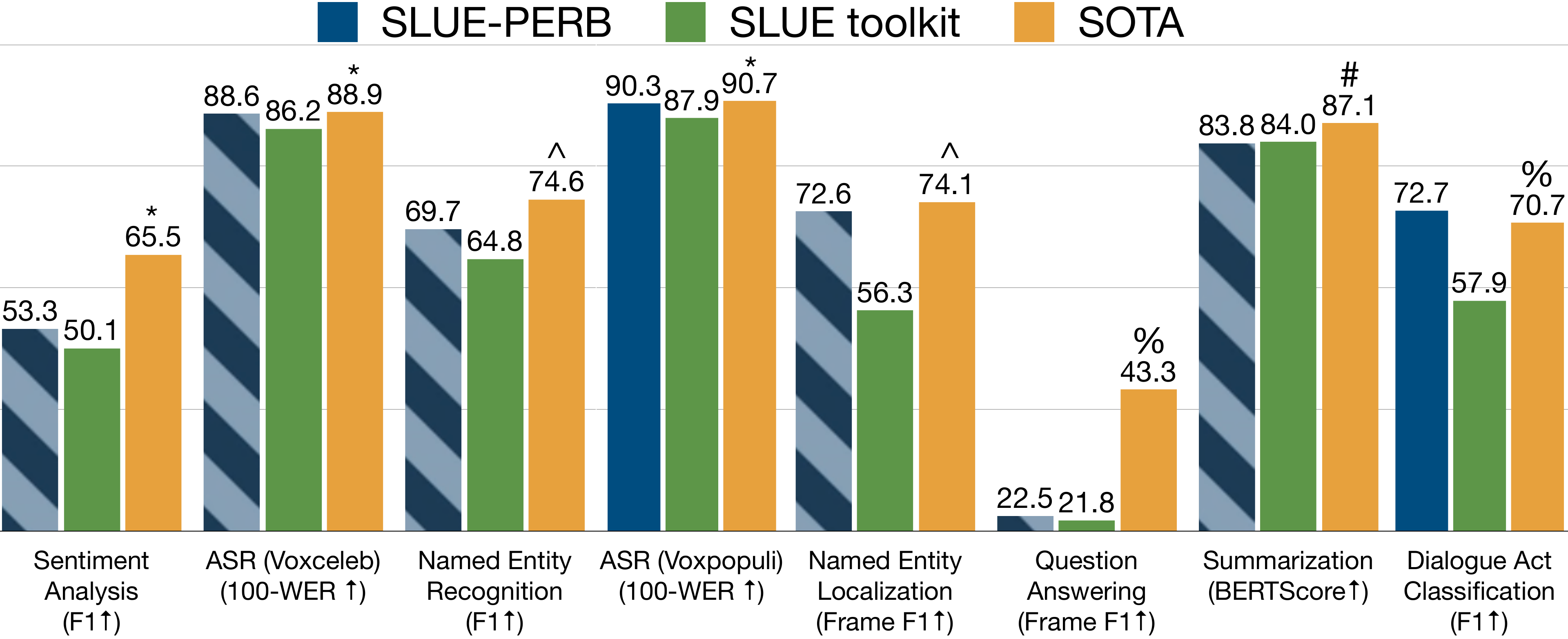}
\vspace{-0.1in}
\caption{Performance of best performing SLUE-PERB results, best E2E model in SLUE toolkit~\cite{shon2022slue,shon2023slue} and SOTA on SLU tasks. SOTA results  from $^*$:\cite{shon2022slue}, $\wedge$:\cite{pasad2022use}, $^\%$:\cite{shon2023slue}, $^\#$:\cite{espnet_summ}. Dense striped bars correspond to the ``complex prediction head'', and solid bars corresponds to ``fine-tuned representations". }
\label{fig:SOTA_result}
\vspace{-0.1in}
\end{figure*}

In this section, we analyze the performance of various \sfms\space on our performance leaderboard SLUE-PERB, as detailed in \sect~3. This analysis provides insights into the types of \sfms\space that prove most effective for complex understanding tasks and how this trend varies across tasks and modeling settings. 
\figs~\ref{fig:SSL_simple}-\ref{fig:SOTA_result} summarize our results.  In all figures, bars with sparse stripes correspond to the ``lightweight prediction head'' setting, dense striped bars correspond to ``complex prediction head'', and solid bars correspond to ``fine-tuned representations".
Test and development set results for all experiments are shown in \tabs~\ref{tab:main-test-results} and \ref{tab:main-dev-results} (in the Appendix) respectively.  
\subsection{Lightweight prediction head}
{\noindent \bf What is the best SSL \sfm\space for SLU?}
We first compare SSL \sfms\space using the ``lightweight prediction head'' evaluation protocol (\sect~\ref{subsec:protocol}) in \fig~\ref{fig:SSL_simple}. We observe that among all SSL models, WavLM features consistently demonstrate superior performance across all tasks, probably since it was pre-trained on larger and more diverse corpora (see \tab~\ref{tab:pre-trained-speech}).
We further observe that HuBERT features outperform Wav2Vec2 on all tasks except DAC. Prior work~\cite{SUPERB} has also noted the superior performance of WavLM and HuBERT's representations.

{\noindent \bf What is the best supervised \sfm\space for SLU?}
\fig~\ref{fig:ASR_simple} compares models that use supervised ASR \sfms\space and are trained with lightweight prediction heads. Our results show that while OWSM is slightly worse than Whisper on SA, NER, and NEL tasks, it significantly outperforms Whisper for DAC and QA. 
As shown in \tab~\ref{tab:pre-trained-speech}, the two models differ in encoder architecture (branchformer in OWSM~\cite{peng2024owsm} vs. transformer in Whisper~\cite{whisper}), training objective (joint Connectionist Temporal Classification (CTC) loss in OWSM~\cite{peng2024owsm}), and pre-training data, which may contribute to the difference in their downstream performance.
Notably, Whisper performs significantly worse on QA. This may result from Whisper's pre-training on 30-second speech segments, while the input audios for QA tasks are typically longer than 30 seconds. While OWSM is also pre-trained on 30 second segments, our results show that Whisper representations particularly struggle to perform well on longer utterances; we discuss this further in Sec.~\ref{subsec:appendix_exp}.
\begin{figure}[t]
\centering
\includegraphics[width=0.98\linewidth]{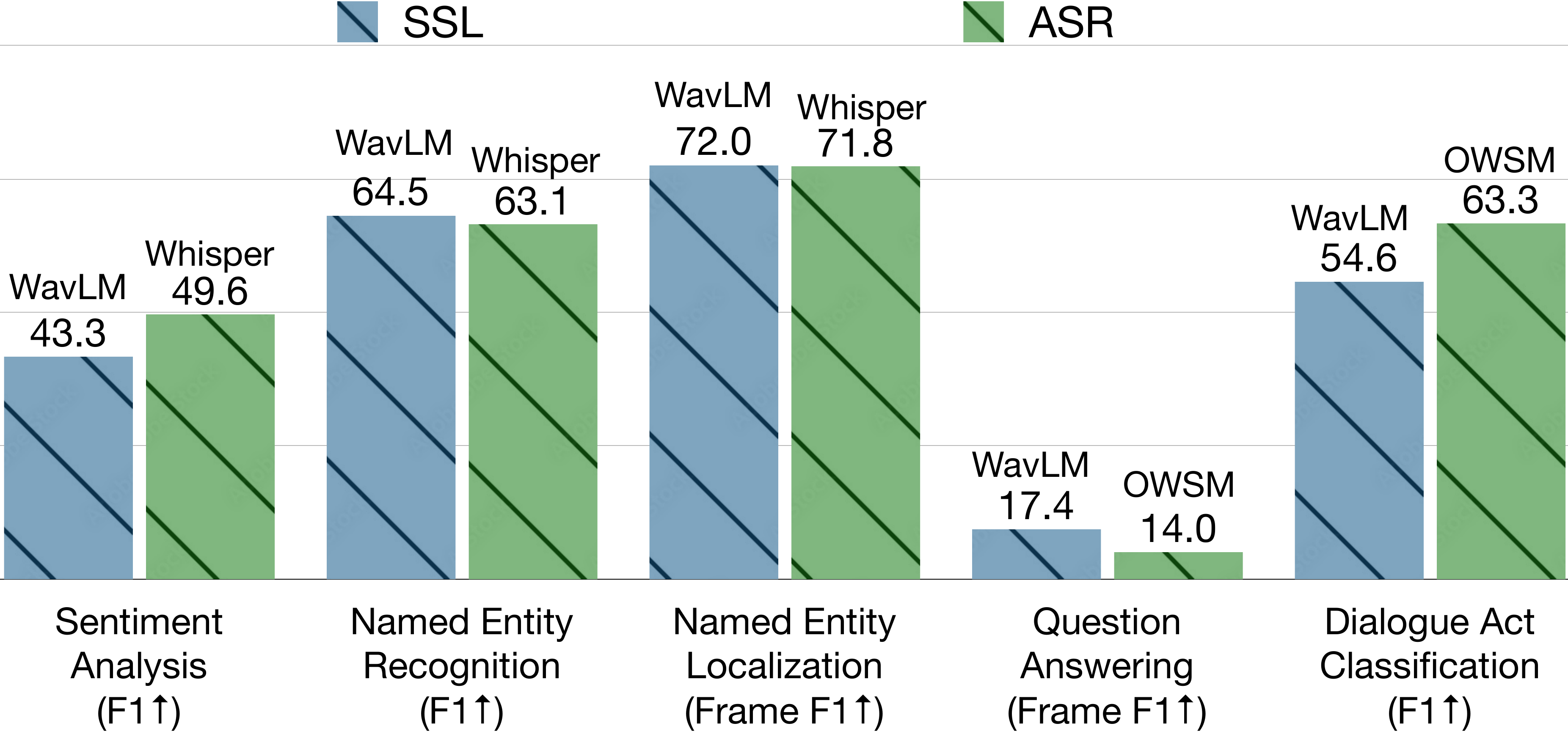}
\vspace{-0.1in}
\caption{Performance of best performing SSL and ASR \sfms\space with a lightweight prediction head on SLUE tasks. The label for each bar is the specific \sfm\space chosen.}
\label{fig:Full_simple}
\vspace{-0.1in}
\end{figure}
\begin{figure}[t]
\centering
\includegraphics[width=0.98\linewidth]{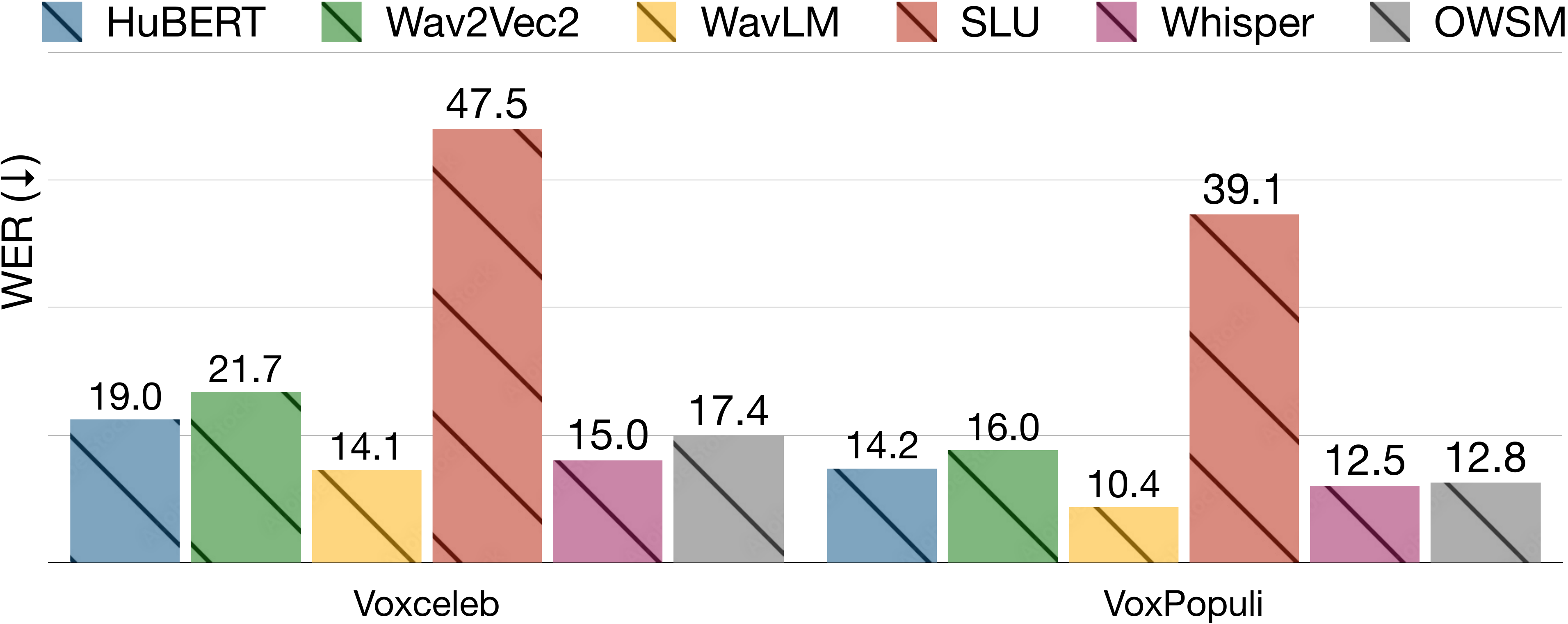}
\vspace{-0.1in}
\caption{ASR performance of \sfms\space with a lightweight prediction head on VoxCeleb and VoxPopuli datasets.}
\label{fig:ASR_results}
\vspace{-0.1in}
\end{figure}

{\noindent \bf SSL vs. supervised \sfms\space for SLU:}
\fig~\ref{fig:Full_simple} reports the performance of the best performing SSL  and ASR \sfms\space using a lightweight prediction head. We can observe that supervised ASR \sfms\space exhibit the best performance on the classification tasks (SA, DAC). Meanwhile, SSL \sfms, WavLM, demonstrate strong performance on temporal alignment and sequence generation tasks, comparable to or better than supervised ASR \sfms. 
Since SSL \sfms \ have an encoder-only architecture, the SLU tasks could leverage all the information learned during pre-training as we use the representations from all encoder layers. Supervised \sfms, on the other hand,
employ an encoder-decoder architecture and may also retain semantic information within their decoder, which is not used for feature extraction in our experiments.
We anticipate that SLU tasks could benefit from integrating the pre-trained decoder of supervised \sfms\space, although we leave this exploration to future work.

Additionally, \tab~\ref{tab:main-test-results} shows that the supervised SLU \sfms\space consistently underperform across all tasks, probably due to their much smaller pre-training data. However, they are comparable to SSL \sfms\space on DAC. This result may be attributed to the scripted nature of conversations in DAC, that resemble the scripted recordings in the SLURP data used for pre-training our SLU model.

We also report the ASR performance for the SLUE Phase-1 datasets in \fig~\ref{fig:ASR_results}. Surprisingly, we observe that features extracted from supervised ASR \sfms\space exhibit worse WER than an SSL \sfm, namely WavLM. As in sequence generation tasks,
we speculate that this may be attributed to
the use of representations from the encoder layers alone.

\subsection{Do performance trends change with different modeling strategies?}
{\noindent \bf Complex prediction head:}
\tab~\ref{tab:main-test-results} and \fig~\ref{fig:Full_complex} show the performance trends of models with a complex prediction head. We observe that the trends remain similar to the setting with simple prediction heads, where WavLM features consistently achieve the best performance across most tasks. Among supervised ASR \sfms, OWSM now outperforms Whisper on most tasks.
SSL \sfms\space demonstrate slight superiority on most temporal alignment and sequence generation tasks, while supervised ASR \sfms\space excel on classification tasks (\fig~\ref{fig:Full_complex}). We note a reduction in the performance gap between different \sfms\space compared to the lightweight prediction head setting. For example, all models now exhibit very similar performance on the SA task. Similarly for SUMM, the performance of all models is very close, but the models that use supervised ASR \sfms\space are slightly better, reinforcing prior work showing the benefits of ASR pre-training for SUMM~\cite{espnet_summ}.

{\noindent \bf Fine-tuned representations:}
Similarly to the trends with frozen representations, \tab~\ref{tab:main-test-results} and \fig~\ref{fig:Full_fine-tune} demonstrate that WavLM features continue to exhibit superior performance among SSL representations, while OWSM performs better than Whisper when we fine-tune \sfms. 
Additionally, \fig~\ref{fig:Full_fine-tune} illustrates that even with complete fine-tuning of \sfms, SSL \sfms \ (WavLM) still performs optimally on sequence generation and temporal alignment tasks, whereas supervised ASR \sfms\space perform better or equally well on classification tasks.

\section{Discussion}
\subsection{Is there an overall best model?}

When comparing the performance between lightweight and complex prediction heads (refer to \figs~\ref{fig:Full_simple} and~\ref{fig:Full_complex}), we notice an improvement in performance across all \sfms\space and tasks. Upon closer examination, it becomes evident that the performance improvement is more pronounced for the SSL \sfms\space compared to supervised ASR \sfms\space on classification tasks, resulting in an overall decrease in the performance gap. 

When comparing performance of frozen and fine-tuned representations under the lightweight prediction head protocol (\figs~\ref{fig:Full_simple} and~\ref{fig:Full_fine-tune}), we generally observe an improvement in performance across all \sfms\space and tasks. However, a notable exception is observed with the supervised ASR \sfms, which perform worse on the NER and NEL tasks. This discrepancy may be attributed to the presence of an excessive number of trainable parameters, especially for the OWSM model, when the entire supervised ASR encoder is fine-tuned.

We further compare the performance achieved by frozen representations with a complex prediction head (Fig.~\ref{fig:Full_complex}) against fine-tuned representations with a lightweight prediction head (Fig.~\ref{fig:Full_fine-tune}).
Interestingly, complex prediction heads demonstrate superior performance compared to fine-tuned representations across most tasks.
However, for the DAC task, fine-tuning a pre-trained encoder yields better results across all \sfms. 

Overall, our findings indicate that there is no \emph{universal} optimal method for incorporating pre-trained SFMs across all tasks.
When we take both \sfms\ and prediction heads into consideration, the optimal \sfms\ and method of incorporating them is task-dependent for our complex SLU tasks.
This is in contrast to some prior works~\cite{SUPERB}, where a single model,  WavLM, emerged as the \emph{universal} best performing model. 

\subsection{Performance-compute tradeoffs}
We also compare the training and inference efficiency of using a complex prediction head and fine-tuned representations, both of which outperform frozen representations with a lightweight prediction head.
Models with a complex prediction head offer overall better performance, as well as greater training efficiency due to their significantly fewer trainable parameters (\tab~\ref{tab:trainable_parameters} in Appendix). However, it's important to note that the use of complex prediction heads leads to a substantial increase in inference time compared to simple prediction heads ($>$ 2.5x for all tasks). In summary, employing a complex prediction head is, in general, better when inference speed is not a bottleneck. On the other hand, if latency is a concern, fine-tuned representations with a lightweight prediction head serve as a good option, enhancing performance without compromising on inference time.

\subsection{Training data tradeoffs} 
The SLUE benchmark comprises datasets with varying amounts of training data, which enables us to consider the effects of both low-resource settings and scenarios where we have sufficient labeled data ($>$ 200 hours).
We observe certain trends in the utility of \sfms\ for lower- vs.~higher-resource tasks in Table 3.
For example, the DAC task, with only 7 hours of training data, has the smallest training set. Supervised \sfms \ appear to be particularly beneficial for DAC across all settings incorporating pre-trained representations. Conversely, for tasks with more abundant labeled data, such as SQA and SUMM, we observe a narrower performance gap between different \sfms. Interestingly, in some cases, SSL \sfms \ like WavLM even outperform supervised \sfms.
We plan to perform a more thorough examination of the effects of varying data size within each task in future work.

\subsection{Comparison with SOTA and E2E baseline}
\fig~\ref{fig:SOTA_result} compares the best results in our SLUE-PERB benchmark with the best E2E results in the original SLUE toolkit~\cite{shon2022slue,shon2023slue} and SOTA results published in prior works. The best performing E2E models in our benchmark either outperform or achieve comparable performance to existing E2E baselines in the SLUE toolkit.
For SA, the SOTA results~\cite{shon2022slue} are obtained by a pipeline consisting of an ASR system, fine-tuned from Wav2Vec2-large, and a NLU system fine-tuned from Deberta-large, on the SLUE-Voxceleb dataset. It is notable that the SOTA results significantly outperform the SLUE-PERB results, likely due to a significantly larger number of trainable parameters (700 million vs. 32.41 million in our best model), as well as stronger semantic processing ability due to the incorporation of a large pre-trained text encoder.
Regarding ASR tasks, we achieve similar performance to SOTA results~\cite{shon2022slue}, and the small performance difference can be attributed to the fact that SOTA models use external language models (LMs) during decoding.

For NER and NEL tasks, the SOTA results~\cite{pasad2022use} perform better than our benchmark models since they leverage external speech and text data to significantly boost performance. There is a significant difference between SOTA results and our best performing benchmark model for QA tasks. The SOTA model~\cite{shon2023slue} is a pipeline system similar to the SOTA SA model. We hypothesize that the performance gap can be attributed to a larger number of trainable parameters (700 million vs. 32.41 million for in best model) as well as the fact that QA is the most semantically challenging among all SLUE tasks and, hence, greatly benefits from incorporating an LM. For SUMM, the SOTA results~\cite{espnet_summ} are achieved by using Whisper-base as the ASR model and a fine-tuned T5-base model for text summarization. The SOTA results outperform our best results, potentially because we do not incorporate a pre-trained LM. We also demonstrate that we outperform the current SOTA~\cite{shon2023slue} on DAC despite having fewer trainable parameters (700M in the SOTA pipeline model vs. 561.91M in our best model).

These findings highlight that the benchmark models are strong baseline E2E models and outperform typical E2E baselines reported in prior work~\cite{shon2022slue,shon2023slue}. By giving open access to these strong baselines as part of SLUE-PERB, we facilitate faster research and development on SLUE tasks. We further show that E2E models can outperform pipeline systems for certain tasks despite having fewer trainable parameters, indicating that the utility of pre-trained LMs is task-dependent. However, pipeline SOTA models currently outperform end-to-end models on semantically challenging SLU tasks like QA and SA. Hence, we plan to extend our benchmark to include pipeline systems in future work to further explore their effectiveness.

\section{Conclusion}
In this paper, we address the lack of performance benchmarks for evaluating pre-trained \sfms\space on SLU tasks. We introduce SLUE-PERB to compare multiple pre-trained SSL and supervised \sfms\space on complex SLU tasks. Our experiments demonstrate that supervised ASR \sfms\space like OWSM produce the best performing representations for classification tasks, while SSL \sfms\space like WavLM can outperform or perform comparably to supervised ASR \sfms\space on temporal alignment and sequence generation tasks. The trends generally remain similar across different evaluation settings, but the performance gap between different \sfms\space decreases as we increase the size of the prediction head or fine-tune the \sfms. We also find that while there is no \emph{universal} best approach for incorporating \sfms, a complex prediction head gives the best performance for most tasks, at the price of higher inference latency.

In addition to providing guidance for researchers working on SLU tasks, we believe that our findings will spark innovation in developing \sfms\ for SLU tasks. (i) SSL \sfm\ representations either outperform or perform comparably to supervised \sfm\ representations for sequence generation tasks. This suggests that supervised \sfms, which employ an encoder-decoder architecture, may retain meaningful information within their decoder, which is not straightforward to use for feature extraction. Hence, developing encoder-only supervised \sfms~\cite{peng2024owsmctc} could be a promising future research direction. (ii) \sfms\ like Whisper demonstrate notably poor performance on QA since Whisper’s pre-training is on 30-second speech segments, while the input audios for QA tasks are typically longer than 30 seconds. This suggests the need for \sfms\ pre-trained on longer speech utterances~\cite{long-hubert}. (iii) Using a complex prediction head with a frozen \sfm\ outperforms full fine-tuning on most tasks, which suggests the exploration of modeling strategies that can utilize \sfms\ without significantly changing the pre-trained parameters, such as using parameter-efficient tuning approaches~\cite{LoRA}.
By making all our code public, we aim to facilitate future research and development on SLUE tasks. 
In future work, we plan to extend SLUE-PERB to include more data and models, including pipeline systems.

\section*{Limitations}
Our approach currently uses only the encoder of the supervised \sfms \ to generate speech representations. A potential limitation is that supervised \sfms \ are encoder-decoder architectures and may also retain some information within their decoder, which is currently not being used in generating speech representations. We plan to delve deeper into generating representations from the pre-trained decoders in future work. \fig~\ref{fig:SOTA_result} also illustrates that pipeline models incorporating large pre-trained text encoders can outperform E2E SLU models on many tasks. Hence, a limitation of our benchmark is that we currently do not include pipeline systems, and we plan to extend our benchmark to incorporate these systems in future work. Further, we observe that full fine-tuning of \sfms \ might be too computationally expensive for some tasks, and we plan to explore the efficacy of parameter-efficient fine-tuning approaches in future work.

\section*{Broader Impact and Ethics}
In this work, we compare various \sfms \ on many complex SLU tasks and gain insights on which \sfms \ perform the best and what is the optimal way of incorporating \sfms \ in E2E SLU systems. 
Our investigations aim to provide valuable insights to researchers regarding which \sfms \ are best suited for their experiments and how to achieve optimal performance with minimal experimentation.
Further, by incorporating \sfms, they can perform the task with a significantly smaller number of trainable parameters and without the need for large amounts of task-specific labeled data. Additionally, we adhere to the ACL Ethics Policy. Our experiments are based on open-source datasets with no violation of privacy, and we will make all our code and models publicly available.

\section*{Acknowledgement}
Experiments in this work used the Bridges2 system at PSC and Delta system at NCSA through allocations CIS210014 and IRI120008P from the Advanced Cyberinfrastructure Coordination Ecosystem: Services \& Support (ACCESS) program, supported by National Science Foundation grants \#2138259,\#tel:2138286, \#tel:2138307, \#tel:2137603, and \#tel:2138296.

\bibliography{custom}
\newpage
\appendix

\section{Appendix}
\label{sec:appendix}
\subsection{Model details}
\label{subsec: model_details}
{\noindent\bf Wav2Vec2}~\cite{wav2vec2} is a SSL speech model which employs a contrastive loss during pre-training and has shown improvements in large-scale ASR. 

{\noindent\bf HuBERT}~\cite{HuBERT} is another SSL model that predicts discrete targets of masked speech regions, similar to the masked language model objective. 

{\noindent\bf WavLM}~\cite{wavlm} expands on HuBERT by increasing pre-training data and adopting a masked speech denoising and prediction framework.

{\noindent\bf Whisper}~\cite{whisper} is one large speech foundation model that has been pre-trained on huge amounts of labeled data for ASR and speech translation (ST) tasks. 

{\noindent\bf OWSM}~\cite{owsm,peng2024owsm} is a reproduction of Whisper using publicly available data and open-source toolkits.
\subsection{Datasets, Tasks and Metrics}
\label{subsec: dataset}
All the datasets are released under Creative Common license to give the best freedom of use.

{\noindent\bf SLUE-VoxCeleb}~\cite{shon2022slue}: SLUE-VoxCeleb is constructed from YouTube videos. 
In this dataset, each spoken utterance is labeled with one of three sentiment classes: positive, negative, and neutral. 
To assess SA performance, we calculate macro-averaged F1 scores.

{\noindent\bf SLUE-Voxpopuli}~\cite{shon2022slue, shon2023slue}: SLUE-Voxpopuli consists of 
European Parliament event recordings.
It includes 7 named-entity tags and 13 sub-tags (fine-grained tagging labels). 
Prior work~\cite{shon2023slue} extends SLUE-VoxPopuli to also evaluate NEL systems by including word-level timestamps for entities.
NEL performance is evaluated either as a frame-level overlap between the predicted and the ground-truth entity spans and is reported as an F1 score ({\it frame-F1}), tuned with an offset hyperparameter~\cite{shon2023slue}. 
The NEL evaluation is purely based on the time stamps and does not consider the entity tags or the entity phrases.
Complementary to NEL, NER performance is evaluated on the predicted named entity phrase and the corresponding tags using a micro-averaged F1 score~\cite{ghannay2018end, shon2022slue}. In addition, we also report {\it label-F1} that only considers the tag predictions and excuses misspellings or segmentation errors in the decoded text. 

{\noindent\bf SLUE-HVB}~\cite{shon2023slue}: HarperValleyBank corpus consists of scripted dialogues between bank employees and customers.
The dialog act labels in SLUE-HVB include 5 actions and 18 sub-actions (fine-grained labeling scheme).
We evaluate DAC on the fine-grained labeling scheme using macro-averaged
(unweighted) F1 score.

{\noindent\bf SLUE-SQA-5}~\cite{shon2023slue}: SLUE-SQA-5 is a spoken question answering (QA) corpus where both document and question consist of real speech data. The question-answer pairs are collected from the text QA dataset; spoken documents are collected from the Spoken Wikipedia dataset~\cite{kohn2016mining} whereas the spoken versions of questions are obtained by crowdsourcing. Similar to NEL, we measure the performance using the frame-F1 score.

{\noindent\bf SLUE-TED}~\cite{shon2023slue}: SLUE-TED is a corpus of summaries for TED-talks. 
The ground truth summary is obtained by concatenating the title and abstract of TED talks, which are publicly available.
We evaluate summarisation performance using ROUGE~\cite{rouge} and BERTScore~\cite{zhang2020bertscore}.

\subsection{Experimental Setups}
\label{subsec:appendix_exp}
All our experiments are conducted with ESPnet-SLU toolkit~\cite{ESPnet-SLU}.
We apply SpecAugment \cite{specaugment} and use dropout \cite{dropout} and label smoothing \cite{label-smoothing} techniques. The models are trained using an NVIDIA A40 (40GB) GPU. All model, training, and inference parameters are selected based on validation performance. Table ~\ref{tab:hp-st-tr} shows training and inference hyperparameters for our hyperparameter search.
We perform extensive tuning of training parameters, particularly warmup and learning rate.
Full details about models, configuration files, and data preparation will be made publicly available prior to publication.

{\noindent\bf Lightweight prediction head:} For classification tasks, the prediction head is a linear classifier that takes in the pooled representations as discussed in Sec.~\ref{sec: experiment}. The output of the classifier layer is the number of classes, which is 3 for SA and 18 for DAC. For NER and NEL, the output is the text transcript, where entity phrases are delimited by entity tags and special characters. An example of NER label sequence is ``we welcome ORG FILL parliament SEP 's agreement'' where ``ORG'' is the entity tag, ``parliament'' is the entity mention, and FILL and SEP are special characters.

For QA, the input is the concatenation of the question and document audio, and the output is the concatenation of the question and document transcript, where special characters again delimit the answer. An example output sequence is ``who is the present quarterback of the broncos SEP nature and persistence of the tennessee volunteers quarterback at the time ANS peyton manning ANS having ...'' where the ``SEP'' token separate the question and document transcript and ``peyton manning'' is the answer to the question delimited by special tokens ``ANS''. Since each spoken document is nearly 40 seconds long, we cannot use Whisper's original sinusoid positional embedding since it cannot accept inputs greater than 30 seconds. Hence, we defined our own sinusoid positional embedding that can accept inputs that are as long as 2 minutes to generate speech representations from the Whisper encoder. Since sinusoid positional embedding does not have any parameters, we believe that our modeling design should not affect the quality of generated speech representations. The architecture of the prediction head for NER and QA are shallow conformer encoders trained with CTC loss, as described in Sec.~\ref{sec: experiment}.

{\noindent\bf Complex prediction head:} The architecture of the complex prediction head is an encoder-decoder architecture consisting of a 12-layer conformer encoder and a 6-layer transformer decoder. For SUMM task, the output is the concatenation of the title and abstract of TED talks, which are publicly available. An example of SUMM label sequence is ``what it's like to be a parent in a war zone $[$sep$]$ how do parents protect their children and help them feel secure again $\cdots$''. Further, for SQA, we obtain the answer tokens from the decoder and then get the timestamps for the answer tokens from greedy CTC decoding. The inference setting for all other non-classification tasks is the same as that with the ``Lightweight prediction head''.

{\noindent\bf Fine-tuned representations:} The architecture of the prediction head is the same as the lightweight prediction head; however, now the pre-trained speech representations are also fine-tuned. Similar to prior work~\cite{wav2vec2, HuBERT, wavlm}, the convolutional feature encoder layers for SSL \sfms \ are kept frozen.

\begin{table*}[t]
  \centering
 {
  \begin{tabular}{lr}
    \toprule
    Hyperparameter & Value \\
    \midrule
    Convolution Subsampling & [1/2x, 1/4x]\\
    Dropout Rate & [0, 0.1, 0.2] \\
    LR schedule & [inv. sqrt., exp. lr.]\\
    Max learning rate & [1e-1, 1e-2, 5e-3, 1e-3, 4e-4, 1e-4, 1e-5, 1e-6] \\
    Warmup steps &  [2500, 5000, 10000] \\
    Number of epochs & [30, 50, 70] \\
    Adam eps  &  1e-8\\
    Adam betas  & (0.9, 0.999)\\
    Weight decay & [1e-5, 1e-6, 1e-7]\\
    \midrule
    Beam Size & [1, 2, 10]\\
    Length Penalty & [0, 0.1]\\
    CTC weight & [0.0, 0.3] \\
    \bottomrule
  \end{tabular}
  }
  \caption{Training and inference hyper-parameter search for SLUE-PERB Models.}
  \label{tab:hp-st-tr}
\end{table*}
\begin{table*}[t]
  \centering
    \resizebox {\linewidth} {!} {
\begin{tabular}{cc|cccccccccc}
\toprule
Evaluation & Pre-Trained &  \multicolumn{2}{c}{SLUE-VoxCeleb}  &  \multicolumn{4}{c}{SLUE-VoxPopuli} & SQA-5 & \multicolumn{2}{c}{SLUE-TED} & SLUE-HVB \\
\cmidrule(r){3-4}\cmidrule(r){5-8}\cmidrule(r){9-9}\cmidrule(r){10-11}\cmidrule(r){12-12}
Protocol &  Model & SA & ASR & \multicolumn{2}{c}{NER} & ASR & NEL & QA & \multicolumn{2}{c}{SUMM} & DAC\\
& & F1 $\uparrow$ & WER $\downarrow$ & Label F1 $\uparrow$ & F1 $\uparrow$ & WER $\downarrow$ & Frame F1 $\uparrow$ & Frame F1 $\uparrow$ & ROUGE-L $\uparrow$ & BERTScore $\uparrow$ & F1 $\uparrow$\\
\midrule
\multirow{2}{*}{Lightweight} & HuBERT (large) & \cellgradsa{37.2} & \cellgradsb{16.2} & \cellgradsc{81.8} & \cellgradsd{64.6} & \cellgradse{13.8} & \cellgradsf{70.9} & \cellgradsg{14.3} & \xmark  & \xmark  & \cellgradsj{46.7}\\
& Wav2Vec2 (large) & \cellgradsa{40.0} & \cellgradsb{18.7} & \cellgradsc{79.9} & \cellgradsd{64.5} & \cellgradse{15.4} & \cellgradsf{68.4} & \hphantom{0}\cellgradsg{6.7} & \xmark  & \xmark  & \cellgradsj{50.6}\\
\multirow{2}{*}{prediction}& WavLM (large) & \cellgradsa{38.9} & \cellgradsb{11.8} & \cellgradsc{87.4} & \cellgradsd{71.4} & \cellgradse{10.2} & \cellgradsf{74.1} & \cellgradsg{18.9} & \xmark  & \xmark  & \cellgradsj{53.5}\\
& Whisper (medium) & \cellgradsa{44.7} & \cellgradsb{13.0} & \cellgradsc{85.8} & \cellgradsd{68.9} & \cellgradse{12.0} & \cellgradsf{73.5} & \hphantom{0}\cellgradsg{0.4} & \xmark  & \xmark  & \cellgradsj{57.2}\\
\multirow{2}{*}{head}& OWSM (3.1) & \cellgradsa{42.2} & \cellgradsb{14.9} & \cellgradsc{84.6} & \cellgradsd{69.2} & \cellgradse{12.6} & \cellgradsf{73.1} & \cellgradsg{15.0} & \xmark & \xmark & \cellgradsj{69.1}\\
& Pre-trained SLU & \cellgradsa{36.6} & \cellgradsb{44.6} & \cellgradsc{66.6} & \cellgradsd{50.8} & \cellgradse{37.7} & \cellgradsf{52.2} & \hphantom{0}\cellgradsg{2.2} & \xmark  & \xmark  & \cellgradsj{56.6}\\
\midrule
\multirow{2}{*}{Complex} & HuBERT (large) & \cellgradsa{46.9} & \cellgradsb{12.8} & \cellgradsc{84.6} & \cellgradsd{69.4} & \cellgradse{12.6} & \cellgradsf{72.7} & \cellgradsg{25.6} & \cellgradsh{16.1} & \cellgradsi{83.4} & \cellgradsj{62.8}\\
& Wav2Vec2 (large) & \cellgradsa{46.5} & \cellgradsb{14.3} & \cellgradsc{83.1} & \cellgradsd{68.9} & \cellgradse{13.1} & \cellgradsf{74.0} & \cellgradsg{22.1} & \cellgradsh{16.3} & \cellgradsi{83.3} & \cellgradsj{67.0}\\
\multirow{2}{*}{prediction}& WavLM (large) & \cellgradsa{47.8} & \hphantom{0}\cellgradsb{9.6} & \cellgradsc{87.9} & \cellgradsd{74.1} & \hphantom{0}\cellgradse{9.5} & \cellgradsf{74.7} & \cellgradsg{25.2} & \cellgradsh{16.7} & \cellgradsi{83.4} & \cellgradsj{70.7}\\
& Whisper (medium) & \cellgradsa{45.2} & \cellgradsb{12.8} & \cellgradsc{86.1} & \cellgradsd{69.9} & \cellgradse{12.7} & \cellgradsf{73.9} &   \cellgradsg{2.0}   & \cellgradsh{16.3} & \cellgradsi{83.7} & \cellgradsj{69.4}\\
\multirow{2}{*}{head}& OWSM (3.1) & \cellgradsa{46.8} & \cellgradsb{14.0} & \cellgradsc{84.8} & \cellgradsd{72.2} & \cellgradse{12.0} & \cellgradsf{70.7} & \cellgradsg{23.7} & \cellgradsh{16.6} & \cellgradsi{83.7} & \cellgradsj{73.5}\\
& Pre-trained SLU & \cellgradsa{45.2} & \cellgradsb{33.5} & \cellgradsc{73.8} & \cellgradsd{61.0} & \cellgradse{27.5} & \cellgradsf{57.8} & \hphantom{0}\cellgradsg{4.2} & \cellgradsh{15.8} & \cellgradsi{83.1} & \cellgradsj{66.8}\\
\midrule
\multirow{3}{*}{Fine-tuning} & HuBERT (large) & \cellgradsa{42.4} & \cellgradsb{12.3} & \cellgradsc{84.3} & \cellgradsd{68.2} & \cellgradse{11.6} & \cellgradsf{73.0} & \xmark & \xmark  & \xmark & \cellgradsj{73.8}\\
& Wav2Vec2 (large) & \cellgradsa{41.8} & \cellgradsb{12.5} & \cellgradsc{84.6} & \cellgradsd{70.4} & \cellgradse{11.3} & \cellgradsf{71.1} & \xmark  & \xmark & \xmark  & \cellgradsj{75.3}\\
& WavLM (large) & \cellgradsa{45.0} & \cellgradsb{10.3} & \cellgradsc{88.3} & \cellgradsd{73.5} & \hphantom{0}\cellgradse{9.3} & \cellgradsf{73.9} & \xmark & \xmark & \xmark & \cellgradsj{75.9}\\
\multirow{3}{*}{representations}& Whisper (medium) & \cellgradsa{48.2} & \cellgradsb{18.2} & \cellgradsc{82.3} & \cellgradsd{65.5} & \cellgradse{16.7} & \cellgradsf{56.3} & \xmark & \xmark & \xmark & \cellgradsj{72.5}\\
& OWSM (3.1) & \cellgradsa{44.2} & \cellgradsb{12.6} & \cellgradsc{83.7} & \cellgradsd{68.3} & \cellgradse{13.7} & \cellgradsf{66.9} & \xmark & \xmark & \xmark & \cellgradsj{76.8}\\
& Pre-trained SLU & \cellgradsa{41.6} & \cellgradsb{31.1} & \cellgradsc{67.5} & \cellgradsd{54.1} & \cellgradse{35.3} & \cellgradsf{54.8} & \xmark & \xmark & \xmark & \cellgradsj{70.3}\\
\bottomrule
\end{tabular}
}
\vspace{-0.1in}
  \caption{Performance of various SSL, supervised ASR, and SLU representations on the development set of SLUE tasks using various evaluation protocols in SLUE-PERB. The symbol \xmark\ indicates that the results were not computed either due to the inability to perform summarization without a decoder or because fine-tuning representations on SQA-5 and SLUE-TED corpora were not feasible within our computational budget.} 
\label{tab:main-dev-results}
\vspace{-0.1in}
\end{table*}

\subsection{Number of Trainable Parameters}

We present the number of trainable parameters for all our models in \tab~\ref{tab:trainable_parameters}. We observe that the lightweight prediction head protocol has approximately 6 million trainable parameters, the complex prediction head setting has around 30 million trainable parameters, and fine-tuning representation has nearly 300 million parameters for most speech representations and tasks. Consequently, the complex prediction head settings serves as a middle ground between lightweight prediction heads and fine-tuned representation settings in terms of computational cost.
Furthermore, we demonstrate that increasing the number of trainable parameters does not always result in improved performance. Interestingly, models with complex prediction heads can outperform models with fine-tuned representations on some SLU tasks, namely NER and NEL. This observation highlights the need to explore diverse methods of incorporating pre-trained speech representations to achieve optimal performance.

\begin{table*}[t]
  \centering
    \resizebox {\linewidth} {!} {
\begin{tabular}{cc|cccccc}
\toprule
Evaluation & Pre-Trained &  \multicolumn{2}{c}{SLUE-VoxCeleb}  &  SLUE-VoxPopuli & SQA-5 & SLUE-TED & SLUE-HVB \\
\cmidrule(r){3-4}\cmidrule(r){5-5}\cmidrule(r){6-6}\cmidrule(r){7-7}\cmidrule(r){8-8}
 Protocol &  Model & SA & ASR & NER & QA & SUMM & DAC\\
\midrule
\multirow{2}{*}{Lightweight} & HuBERT (large) & \hphantom{00}1.1 & \hphantom{00}6.5 & \hphantom{00}6.5 & \hphantom{0}9.7 & \xmark & \hphantom{00}1.1\\
& Wav2Vec2 (large) & \hphantom{00}1.1 & \hphantom{00}6.5 & \hphantom{00}6.5 & \hphantom{0}9.7 & \xmark & \hphantom{00}1.1\\
\multirow{2}{*}{prediction}& WavLM (large) & \hphantom{00}1.1 & \hphantom{00}6.5 & \hphantom{00}6.5 & \hphantom{0}9.7 & \xmark & \hphantom{00}1.1\\
& Pre-trained SLU & \hphantom{00}0.3 & \hphantom{00}9.1 & \hphantom{00}9.1 & 12.2 & \xmark & \hphantom{00}0.3\\
\multirow{2}{*}{head}& Whisper (medium) & \hphantom{00}1.1 & \hphantom{00}9.1 & \hphantom{00}9.1 & \hphantom{0}9.7 & \xmark & \hphantom{00}1.1\\
& OWSM (3.1) & \hphantom{00}1.1 & \hphantom{00}9.1 & \hphantom{00}9.1 & 12.3 & \xmark & \hphantom{00}1.1\\\midrule
\multirow{2}{*}{Complex} & HuBERT (large) & \hphantom{0}32.4 & \hphantom{0}32.4 & \hphantom{0}32.4 & 32.4 & 31.9 & 114.3\\
& Wav2Vec2 (large) & \hphantom{0}32.4 & \hphantom{0}32.4 & \hphantom{0}32.4 & 32.4 & 31.9 & 114.3\\
\multirow{2}{*}{prediction}& WavLM (large) & \hphantom{0}32.4 & \hphantom{0}32.4 & \hphantom{0}32.4 & 32.4 & 31.9 & 114.3\\
& Pre-trained SLU & \hphantom{0}34.9 & \hphantom{0}34.9 & \hphantom{0}34.9 & 34.9 & 34.4 & 124.5\\
\multirow{2}{*}{head}& Whisper (medium) & \hphantom{0}32.4 & \hphantom{0}32.4 & \hphantom{0}32.4 & 32.4 & 31.9 & 114.3\\
& OWSM (3.1) & \hphantom{0}32.4 & \hphantom{0}32.4 & \hphantom{0}35.0 & 35.0 & 34.5 & 124.5\\\midrule
\multirow{3}{*}{Fine-tuning} & HuBERT (large) & 313.4 & 318.9 & 318.9 & \xmark & \xmark & 313.5\\
& Wav2Vec2 (large) & 314.2 & 319.7 & 319.7 & \xmark & \xmark & 314.3\\
& WavLM (large) & 312.3 & 317.8 & 317.8 & \xmark & \xmark & 312.3\\
\multirow{3}{*}{representations}& Pre-trained SLU & \hphantom{0}83.5 & \hphantom{0}93.3 & \hphantom{0}92.3 & \xmark & \xmark & \hphantom{0}83.5\\
& Whisper (medium) & 306.7 & 314.8 & 314.8 & \xmark & \xmark & 306.8\\
& OWSM (3.1) & 561.9 & 569.9 & 569.9 & \xmark & \xmark & 561.9\\
\bottomrule
\end{tabular}
}
  \caption{Number of trainable parameters (in million of parameters) in models using various SSL, supervised ASR, and SLU representations across different evaluation protocols in SLUE-PERB. The symbol \xmark\ indicates that the results were not computed either due to the inability to perform summarization without a decoder or because fine-tuning representations on SQA-5 and SLUE-TED corpora were not feasible within our computational budget. } 
\label{tab:trainable_parameters}

\end{table*}
\end{document}